\pdfoutput=1
\documentclass[10pt,twocolumn,letterpaper]{article}

\usepackage{cvpr}
\usepackage{multirow}
\usepackage[accsupp]{axessibility}

\newcommand{\Fig}[1]{Figure~\ref{fig:#1}}

\newcommand{\Eq}[1]{Eq.~(\ref{eq:#1})}
\newcommand{\Tbl}[1]{Table~\ref{tab:#1}}

\def\ie{\textit{i.e.}}
\def\eg{\textit{e.g.}}

\definecolor{brown}{rgb}{0.85, 0.15, 0.15}
\definecolor{purp}{rgb}{0.95, 0.16, 0.65}
\definecolor{purpc}{rgb}{0.95, 0.36, 0.65}
\definecolor{darkpurple}{rgb}{0.54, 0.17, 0.89}
\definecolor{orange}{rgb}{0.9, 0.45, 0.0}
\definecolor{blue}{rgb}{0.0, 0.5, 1.0}
\definecolor{green}{rgb}{0, 0.8, 0}
\definecolor{darkgreen}{rgb}{0, 0.6, 0}
\definecolor{lgreen}{rgb}{0.6, 0.8, 0}
\definecolor{red}{rgb}{0.8, 0, 0}
\definecolor{redd}{rgb}{0.9, 0, 0}
\definecolor{yellow}{rgb}{0.75, 0.56, 0}
\definecolor{darkblue}{rgb}{0.2, 0.2, 0.8}
\definecolor{brinkpink}{rgb}{0.98, 0.38, 0.5}
\definecolor{cadmiumred}{rgb}{0.89, 0.0, 0.13}
\definecolor{ceruleanblue}{rgb}{0.16, 0.32, 0.75}
\definecolor{dandelion}{rgb}{0.94, 0.88, 0.19}
\definecolor{bostonuniversityred}{rgb}{0.8, 0.0, 0.0}
\definecolor{brown(web)}{rgb}{0.65, 0.16, 0.16}
\definecolor{cornellred}{rgb}{0.7, 0.11, 0.11}
\definecolor{greend}{rgb}{0.0, 0.35, 0.0}

\newcommand{\cmark}{\ding{51}}

\definecolor{err}{rgb}{0.99, 0, 0}

\definecolor{grey}{rgb}{0.9, 0.9, 0.9}

\newcommand{\hyperfootnote}[1][]{\def\ArgI\hyperfootnoteRelay}
\newcommand\hyperfootnoteRelay[2][]{\href{#1#2}{\ArgI}\footnote{\href{#1#2}{#2}}}

\usepackage{enumitem}

\usepackage{makecell}
\usepackage{bbm}
\usepackage{pifont}

\usepackage{xcolor}
\usepackage{soul}

\usepackage{algorithm}
\usepackage{algpseudocode}
\definecolor{cvprblue}{rgb}{0.21,0.49,0.74}
\usepackage[pagebackref,breaklinks,colorlinks,allcolors=cvprblue]{hyperref}

\def\paperID{32236}
\def\confName{CVPR}
\def\confYear{2026}

\title{Robust Promptable Video Object Segmentation}

\author{%
  Sohyun Lee$^{1}$ \hspace{3mm} Yeho Gwon$^{1}$ \hspace{3mm} Lukas Hoyer$^{2}$ \hspace{3mm} Konrad Schindler$^{3}$ \hspace{3mm} Christos Sakaridis$^{3}$ \hspace{3mm} Suha Kwak$^{1}$ \vspace{3mm}\\
    $^1$POSTECH \qquad \ \
    $^2$Google \qquad \ \
    $^3$ETH Zürich  \\
{\tt \small \url{https://sohyun-l.github.io/RobustPVOS_project_page/}}
  }

\begin{document}
\maketitle
\begin{abstract}
The performance of promptable video object segmentation (PVOS) models substantially degrades under input corruptions, which prevents PVOS deployment in safety-critical domains. This paper offers the first comprehensive study on robust PVOS (RobustPVOS). We first construct a new, comprehensive benchmark with two real-world evaluation datasets of 351 video clips and more than 2,500 object masks under real-world adverse conditions. At the same time, we generate synthetic training data by applying diverse and temporally varying corruptions to existing VOS datasets. Moreover, we present a new RobustPVOS method, dubbed Memory-object-conditioned Gated-rank Adaptation (MoGA). The key to successfully performing RobustPVOS is two-fold: effectively handling object-specific degradation and ensuring temporal consistency in predictions. MoGA leverages object-specific representations maintained in memory across frames to condition the robustification process, which allows the model to handle each tracked object differently in a temporally consistent way. Extensive experiments on our benchmark validate MoGA's efficacy, showing consistent and significant improvements across diverse corruption types on both synthetic and real-world datasets, establishing a strong baseline for future RobustPVOS research. Our benchmark is publicly available at \href{https://sohyun-l.github.io/RobustPVOS_project_page/}{our project page}.
\end{abstract}

\section{Introduction}
\label{sec:intro}

Promptable video object segmentation (PVOS) has emerged as a powerful paradigm, enabling users to segment and track arbitrary objects in videos given free-form prompts~\cite{ravi2025sam,Mei_2025_CVPR,Li_2024_CVPR,ding2025sam2long,xiong2025efficient}.
The recent introduction of SAM2~\cite{ravi2025sam} marks a significant breakthrough in this field, achieving impressive zero-shot performance by extending the success of promptable segmentation from images to videos.
However, we empirically find that the performance of SAM2 substantially degrades under input corruptions caused by noise, blur, low illumination, and adverse weather.
This lack of robustness poses critical challenges for deploying PVOS in safety-critical domains such as autonomous vehicles and robotics, where adverse conditions are inevitable, but has not been adequately studied or addressed to date.

\begin{figure}[t!]
    \centering
    \includegraphics[width=\linewidth]{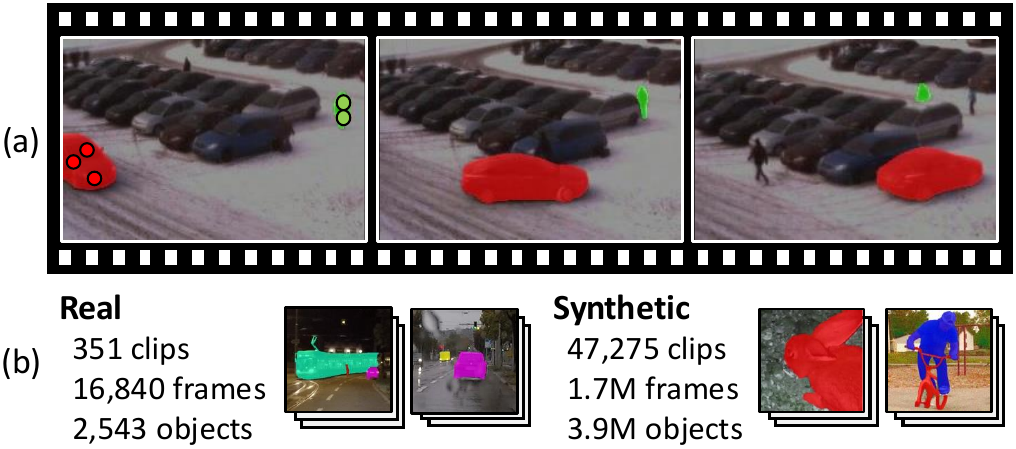}
    \caption{
    Overview of (a) RobustPVOS and (b) our benchmark. RobustPVOS is the task of tracking and segmenting objects, indicated by initial prompts, across frames despite adverse conditions.
    }
    \label{fig:teaser}
\end{figure}
Motivated by this, we introduce the first comprehensive study on robust promptable video object segmentation (RobustPVOS), whose goal and benchmark composition are illustrated in \Fig{teaser}.
Our primary contribution is establishing RobustPVOS as a well-defined research direction through a carefully designed benchmark that enables development and systematic evaluation of RobustPVOS models.
We construct two real-world evaluation datasets comprising 351 video clips with over 2,500 annotated object masks from existing video collections captured under natural adverse conditions including rain, fog, snow, and nighttime scenarios~\cite{acdc,acdc:v2,ji2023mvss}.
To our knowledge, this is the first effort to provide dense, object-level annotations for promptable video segmentation under real-world corruptions.
Additionally, we create synthetic training data by applying eight diverse corruption types with temporal variations~\cite{Hendrycks2019_ImageNetC, chen2024robustsam,kim2023exploring} to annotated videos from established VOS datasets~\cite{MOSE,xu2018youtube,Perazzi_CVPR_2016}, enabling model training and evaluation under controlled conditions.
These datasets provide a valuable testbed for assessing model performance across diverse degradation scenarios, from controlled synthetic corruptions to naturally occurring adverse conditions.

Beyond the benchmark, we further investigate methodological approaches for RobustPVOS.
One potential approach to RobustPVOS is applying existing robust image segmentation methods~\cite{son2020urie,lee2025garasam,AirNet,Bi_2024_CVPR,chen2024robustsam} to individual video frames.
While these methods effectively handle corrupted images, they face fundamental limitations in PVOS, since real-world videos exhibit complex degradation patterns that vary both spatially and temporally: different objects within the same scene are affected differently
(\eg, in fog, distant objects are barely visible while close objects remain clearly visible),
and degradation patterns change across frames.
Robust image segmentation models that process frames independently often yield temporally inconsistent segmentation results.
Also, since they process the entire area of each frame uniformly for improving robustness, they fail to consider the varying effects of corruptions on different objects within the same scene.

To address these challenges, we propose {Memory-object-conditioned Gated-rank Adaptation}, dubbed MoGA, as an effective baseline method for RobustPVOS.
Our key insight is that object-specific representations maintained across frames, which are inherent in modern video segmentation models~\cite{ravi2025sam,Mei_2025_CVPR,zhou2024rmem}, naturally capture how each object is uniquely affected by degradations over time.
By conditioning the robustification process
on the representations of these objects in the memory, MoGA enables the model to handle each tracked object differently while maintaining temporal consistency.
This design demonstrates that memory-based conditioning can substantially improve robustness in PVOS.

Extensive experiments on our benchmark validate the effectiveness of our approach. MoGA achieves consistent improvements across diverse corruption types on both synthetic and real-world datasets, while maintaining competitive performance on clear videos.
Our benchmark and baseline together establish foundations for future research in RobustPVOS.
In summary, our contribution is three-fold:
\begin{itemize}
    \item We initiate the study of RobustPVOS, establishing it as a critical research direction for real-world video understanding applications.
    \item We construct a comprehensive RobustPVOS benchmark, including real-world evaluation datasets with object-level annotations under adverse conditions and synthetic training data with controlled temporal degradations.
    \item We present MoGA as an effective baseline method that addresses the unique challenges of RobustPVOS through object-specific and temporally consistent robustification.
\end{itemize}

\section{Related Work}

\subsection{Video Object Segmentation}
Video object segmentation (VOS) aims to identify and track target objects throughout video sequences~\cite{ravi2025sam}. Traditional methods rely on dense optical flow~\cite{tsai2016video,Fedynyak_2024_WACV,tokmakov2017learning_mem,tokmakov2017learning_motion} to maintain temporal consistency.
Recent approaches leverage memory modules to store object representations over videos~\cite{zhou2024rmem,Mei_2025_CVPR,ravi2025sam,song2020kernelized}, enabling stable tracking.
RMem~\cite{zhou2024rmem} investigates the effect of a restricted memory bank, while SAM-I2V~\cite{Mei_2025_CVPR} focuses on exploiting pre-trained SAM by introducing a memory prompt generator. Notably, Cutie~\cite{cheng2024putting} employs object queries to construct an object-level memory in a top-down manner.
As PVOS has emerged as a new paradigm for prompt-based object segmentation~\cite{ravi2025sam,Mei_2025_CVPR,guo2024x}, UniVS~\cite{Li_2024_CVPR} accepts visual and text prompts and segments objects by averaging their features.
However, these methods are primarily designed for clear videos and struggle under adverse conditions~\cite{li2024learning}, motivating our work on RobustPVOS.

\subsection{Promptable Segmentation}
The Segment Anything Model (SAM)~\cite{kirillov2023segment} revolutionized image segmentation by enabling interactive mask generation.
SAM accepts diverse prompt types such as points, boxes, and masks, demonstrating remarkable zero-shot generalization across domains.
HQ-SAM~\cite{sam_hq} extends SAM for high-quality segmentation by introducing a learnable output token with intricate structures.
SAM2~\cite{ravi2025sam} extends the promptable segmentation ability of SAM to videos by introducing a streaming memory architecture that propagates the masks of prompted objects across frames.
SEEM~\cite{zou2023segment} and COSINE~\cite{liu2025unified} support diverse prompt types, including text and images, through a unified decoding mechanism with learnable memory prompts.
CLIPSeg~\cite{lueddecke22_cvpr} pioneers text-based segmentation by combining CLIP with a transformer decoder for both text queries and visual examples, while REM~\cite{bagchi2025refereverything} adopts a text-to-image generative diffusion model to segment a moving object in a video.
While these methods excel on clear data, their robustness under corruptions remains unexplored, particularly for video applications where temporal consistency is crucial.

\begin{figure*}[t!]
\centering
    \includegraphics[width=1.0\linewidth]{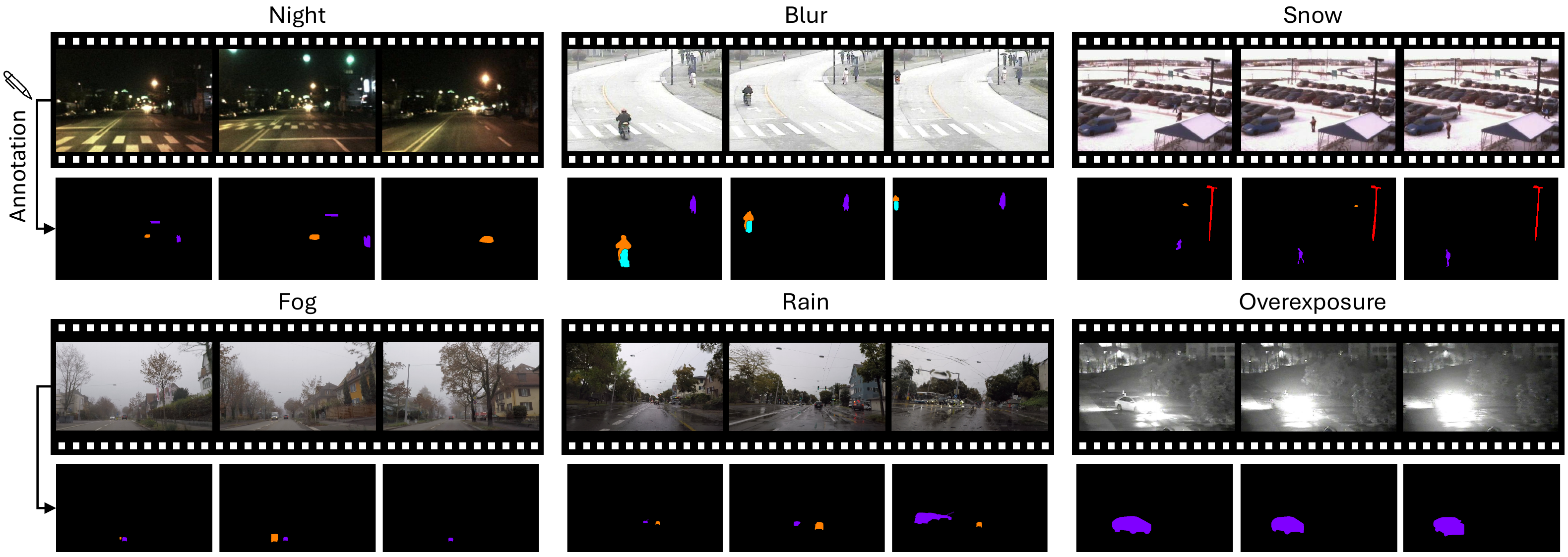}
    \caption{
    Example images and annotated object masks of the real-world evaluation dataset.
    } \label{fig:overview_dataset_real}
\end{figure*}
\subsection{Robustness}
Robustness to input corruptions has been extensively studied for the image domain~\cite{lee2022fifo,lee2023human,Bi_2024_CVPR,lee2024frest,chen2024robustsam,kirillov2023segment,lee2025testdg,lee2025garasam}.
For semantic segmentation, recent works propose various strategies, such as FIFO~\cite{lee2022fifo}, which learns fog-invariant features, and FreD~\cite{Bi_2024_CVPR}, which employs frequency-domain analysis.
RobustSAM~\cite{chen2024robustsam} improves the robustness of SAM~\cite{kirillov2023segment} by introducing anti-degradation modules.
GaRA-SAM~\cite{lee2025garasam} introduces gated-rank adaptation, achieving input-adaptive robustness for SAM through selective (rank-1) component activation.
Universal restoration methods~\cite{son2020urie,AirNet,potlapalli2023promptir,ai2024lora} restore corrupted images and show that restoration improves robust recognition.
However, they process frames independently, so they cannot model temporal consistency in videos.
For robustness in the video domain, VPSeg~\cite{guo2024vanishing} exploits vanishing point priors for robust video semantic segmentation in driving scenes.
Event cameras are also used to better understand motion in videos~\cite{li2024event,kim2022event}, particularly in challenging conditions such as motion blur and low illumination.
Although they handle the temporal dynamics of videos, they still fall short in RobustPVOS due to its multi-object nature.
Our work fills this gap by introducing memory-based conditioning, specifically designed for temporally consistent and object-specific robustification.

\section{RobustPVOS Benchmark}
\label{sec:benchmark}

\begin{table}[t]
  \centering
  \caption{Statistics of our benchmark for RobustPVOS. The top block lists real-world evaluation sets; the bottom blocks summarize the synthetic training set and the synthetic evaluation sets.}
  \label{tab:dataset-stats}
  \resizebox{0.48\textwidth}{!}{
  \begin{tabular}{lrrr}
    \toprule
    Dataset & Clips & Frames & Objects \\
    \midrule
    \multicolumn{4}{l}{\textbf{Real-world evaluation set}} \\
    \addlinespace[2pt]
    ACDC-Video & 149 & 3{,}259 & 613 \\
    MVSeg      & 202 & 13{,}581 & 1{,}930 \\
    \midrule
    \multicolumn{4}{l}{\textbf{Synthetic training set}} \\
    \addlinespace[2pt]
    \makecell[l]{MOSE-C + DAVIS-C\\+ YouTube-VOS-C} & 38{,}552 &  1{,}421{,}464 & 2{,}791{,}712 \\
    \midrule
    \multicolumn{4}{l}{\textbf{Synthetic evaluation set}} \\
    YouTube-VOS-C& 507 & 13{,}710 & 25,574 \\
    \bottomrule
  \end{tabular}}
\end{table}

We present the first RobustPVOS benchmark suite that includes (1) manually annotated real-world evaluation datasets under adverse conditions and (2) a synthetic corruption dataset with temporally varying degradation patterns.
The overall composition of our benchmark is summarized in \Tbl{dataset-stats}, and qualitative examples of the real-world evaluation dataset are presented in \Fig{overview_dataset_real}.

\paragraph{Real-World Evaluation Datasets.}
We curate two real-world PVOS test sets by collecting videos captured under natural adverse conditions from the ACDC-Video~\cite{acdc:v2} and MVSeg~\cite{ji2023mvss} datasets.
While these datasets provide masks, they were not originally designed for PVOS: MVSeg provides only class-level masks, and ACDC offers instance-level masks without temporally consistent tracking.
We therefore annotate dense, per-object, pixel-level masks with consistent instance identities across frames, labeling instance IDs and correcting low-quality masks.
We also manually excluded clips with mild adverse conditions or static objects to ensure suitability for RobustPVOS evaluation.
As shown in Table~\ref{tab:dataset-stats}, we collect 149 clips (3,259 frames) captured in fog, snow, rain, and nighttime conditions from ACDC-Video~\cite{acdc:v2}, annotating 613 object masks on key frames.
From MVSeg~\cite{ji2023mvss}, we annotate 202 video clips (13,581 frames) across various conditions and release them as a public PVOS dataset, providing 1,930 annotated object masks on key frames (Table~\ref{tab:dataset-stats}); for RobustPVOS evaluation, we further construct a dedicated test subset, dubbed MVSeg-adv, by retaining only the clips captured under challenging conditions such as low light, rain, snow, noise, and motion blur, excluding those with mild adverse conditions or static objects.

\begin{figure*}[t!]
    \centering
    \includegraphics[width=\linewidth]{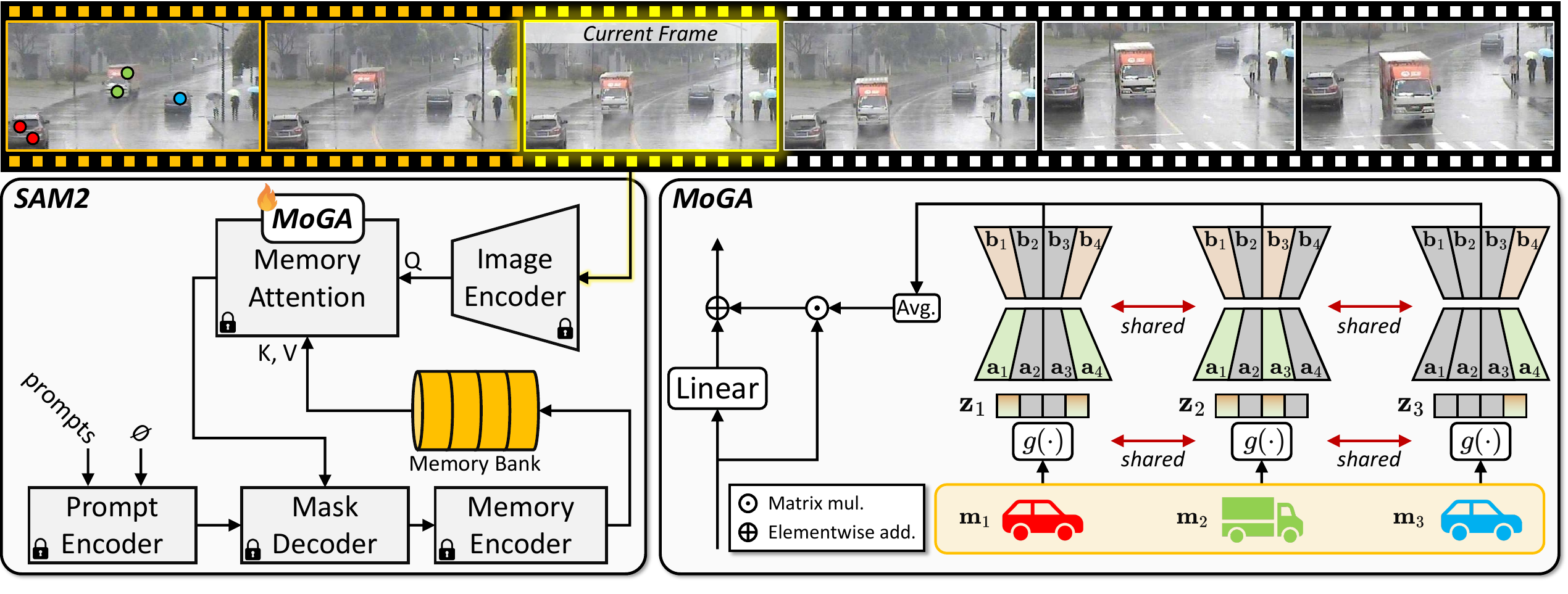}
    \caption{Overview of
    MoGA integrated into SAM2~\cite{ravi2025sam}.
    \emph{Top}: an example input video under adverse weather conditions. Frames with orange outlines are previously processed and stored in the memory bank, while the yellow outline indicates the current frame.
    \emph{Bottom left}: MoGA modules are integrated into the memory attention mechanism of SAM2.
    \emph{Bottom right}: our memory-conditioned gating mechanism where object pointers from the memory bank guide the selection of (rank-1) components of the shared adapter, enabling object-specific robustness while maintaining temporal consistency. Best viewed in color.}
    \label{fig:architecture}
\end{figure*}

\paragraph{Synthetic Corruption Dataset.}
For both training and controlled evaluation, we further construct a synthetic dataset by applying eight types of corruption (\ie, color jitter, Gaussian noise, ISO noise, motion blur, resampling blur, fog, rain, and snow) to existing VOS datasets.
Using corruption synthesis algorithms~\cite{Hendrycks2019_ImageNetC, chen2024robustsam} with Fourier-based temporal modulation~\cite{kim2023exploring}, we vary corruption intensity smoothly across frames to simulate degradation patterns in real videos.
For training, we corrupt videos from MOSE~\cite{MOSE}, YouTube-VOS~\cite{xu2018youtube}, and DAVIS~\cite{Perazzi_CVPR_2016}, generating 46,768 clips (1,774,560 frames) with 3,872,048 object masks.
For evaluation, we create YouTube-VOS-C, a test set containing 507 corrupted clips (13,710 frames) with 25,574 object masks from the YouTube-VOS validation set.

\paragraph{Benchmarking Protocol.}
For each benchmark video, we provide prompts for up to three annotated objects per clip, following the training protocol.
Models must segment and track these objects throughout the entire sequence.
Their performance is evaluated using the standard VOS metrics~\cite{xu2018youtube,ravi2025sam}: region similarity ($\mathcal{J}$), contour accuracy ($\mathcal{F}$), and their combination ($\mathcal{J}\&\mathcal{F}$).
Scores are averaged over objects and then over sequences; for ACDC-Video, whose classes are heavily imbalanced, we average over semantic classes instead.

\section{Method}

We present Memory-object-conditioned Gated-rank Adaptation (MoGA) for RobustPVOS.
The overall architecture of MoGA is illustrated in Figure~\ref{fig:architecture}.
The remainder of this section provides
preliminaries on low-rank adaptation in Section~\ref{sec:lora}
and elaborates on MoGA
in Section~\ref{sec:moga}.

\subsection{Preliminaries on Low-Rank Adaptation} \label{sec:lora}
Low-rank Adaptation (LoRA)~\cite{hu2022lora} is a parameter-efficient fine-tuning technique that freezes a pre-trained weight matrix $\mathbf{W}_0 \in \mathbb{R}^{D \times K}$ and introduces a learnable low-rank adapter $\Delta \mathbf{W} = \mathbf{B}\mathbf{A}$, where $\mathbf{B} \in \mathbb{R}^{D \times R}$ and $\mathbf{A} \in \mathbb{R}^{R \times K}$ with rank $R \ll \min(D, K)$.
This adapter modifies the forward pass as:
\begin{equation}
    \mathbf{h} = \mathbf{W}_0 \mathbf{x} + \mathbf{B}\mathbf{A}\mathbf{x},
\end{equation}
where $\mathbf{x} \in \mathbb{R}^K$ is the input and $\mathbf{h} \in \mathbb{R}^D$ is the output.
Recent work~\cite{lee2025garasam} has shown that decomposing a low-rank adapter into components and selectively activating them can improve robustness to input degradation, but it does not guarantee object-wise temporally consistent robustification.

\subsection{MoGA}
\label{sec:moga}

For RobustPVOS, MoGA leverages object representations maintained across frames in modern video segmentation architectures~\cite{ravi2025sam,Mei_2025_CVPR,zhou2024rmem}.
These architectures employ a memory bank as a temporal storage that accumulates object features from processed frames to enable consistent tracking.
MoGA decomposes the weight matrix of the low-rank adapter, \ie, $\Delta \mathbf{W}$ of Section~\ref{sec:lora}, into (rank-1) components,
and selectively activates them using object-specific representations from the memory bank that stores information of the object from previous frames.
This design enables the adapter to handle distinct objects differently while ensuring the temporal consistency.

\paragraph{(Rank-1) Component Decomposition.}
Following previous work~\cite{lee2025garasam}, we decompose the learnable weight matrix $\Delta \mathbf{W} = \mathbf{B}\mathbf{A}$ into $R$ (rank-1) components $\{ \mathbf{a}_i, \mathbf{b}_i \}_{i=1}^{R}$, where $\mathbf{a}_i \in \mathbb{R}^K$, $\mathbf{b}_i \in \mathbb{R}^{D}$, and $\Delta \mathbf{W} = \sum_{i=1}^{R} \mathbf{b}_i \mathbf{a}_i^\top$.
This decomposition enables a flexible construction of the weight matrix based on the input characteristics.

\begin{table*}[t]
    \centering
    \caption{Evaluation results on real-world and synthetic datasets. We evaluate on MVSeg, ACDC-Video, and YouTube-VOS-C and its counterpart.}
    \begin{tabular}{l|cccccccccccc}
        \toprule
        \multirow{2}{*}{Method} 
        & \multicolumn{3}{c}{\textbf{MVSeg-adv}~\cite{ji2023mvss}} 
        & \multicolumn{3}{c}{\textbf{ACDC-Video}~\cite{acdc:v2}} 
        & \multicolumn{3}{c}{\textbf{YouTube-VOS-C}} 
        & \multicolumn{3}{c}{\textbf{YouTube-VOS}~\cite{xu2018youtube}} \\
        \cmidrule(lr){2-4} \cmidrule(lr){5-7} \cmidrule(lr){8-10} \cmidrule(lr){11-13}
        & $\mathcal J \& \mathcal F$ & $\mathcal J$ & $\mathcal F$ 
        & $\mathcal J \& \mathcal F$ & $\mathcal J$ & $\mathcal F$ 
        & $\mathcal J \& \mathcal F$ & $\mathcal J$ & $\mathcal F$ 
        & $\mathcal J \& \mathcal F$ & $\mathcal J$ & $\mathcal F$ \\
        \midrule
        SAM2~\cite{ravi2025sam} 
            & 69.6 & 60.5 & 78.6 
            & 63.5 & 49.8 & 77.2 
            & 78.7 & 77.2 & 80.2 
            & 82.2 & 80.4 & 83.9 \\
        URIE~\cite{son2020urie}+SAM2 
            & 69.6 & 60.7 & 78.4 
            & 60.9 & 46.4 & 75.3 
            & 78.6 & 77.1 & 80.2 
            & - & - & - \\
        AirNet~\cite{AirNet}+SAM2 
            & 69.1 & 60.4 & 77.7 
            & 59.8 & 46.2 & 73.5 
            & 78.6 & 77.1 & 80.0 
            & - & - & - \\
        GaRA~\cite{lee2025garasam}+SAM2 
            & 69.7 & 61.4 & 78.0 
            & 61.3 & 47.4 & 75.2 
            & 78.1 & 76.3 & 79.9 
            & 79.8 & 78.1 & 81.4 \\
        MoGA+SAM2 
            & \textbf{71.8} & \textbf{62.9} & \textbf{80.7} 
            & \textbf{64.5} & \textbf{50.3} & \textbf{78.8} 
            & \textbf{79.9} & \textbf{78.3} & \textbf{81.5} 
            & \textbf{82.6} & \textbf{80.7} & \textbf{84.4} \\
        \bottomrule
    \end{tabular}\label{tab:main_results}
\end{table*}

\paragraph{Memory-object-conditioned Gating.}
The key innovation in MoGA is conditioning the gating mechanism on
object information stored in the memory bank.
SAM2~\cite{ravi2025sam} maintains object pointers $\mathcal{M} = \{\mathbf m_o\}_{o=1}^{O}$ in the memory bank which accumulate information across frames.
Each object pointer $\mathbf{m}_o \in \mathbb{R}^{d}$ encodes the historical characteristics specific to object $o$.
We introduce a gating module $g(\cdot)$ that takes an object pointer as input and computes object-specific binary gating masks $\textbf{z}_o=g(\mathbf{m}_o) \in \{0, 1\}^R$.
The gating module, which consists of a three-layer MLP followed by Gumbel-Sigmoid relaxation, is shared but applied per-object using its corresponding object pointer $\mathbf m_o$:
\begin{equation}
    \boldsymbol\alpha_{o} = \textrm{MLP}(\textbf{m}_o)
\end{equation}
To enable differentiable learning with discrete selection, we apply Gumbel-sigmoid relaxation~\cite{jang2017categorical} to the logit vectors:
\begin{equation}
    \label{eq:logit}
    \tilde{z}_{o,i} = \sigma \left( \frac{1}{\tau} (\alpha_{o,i} + G_i) \right), \quad i = 1, \ldots, R,
\end{equation}
where $G_i \sim \text{Gumbel}(0,1)$, $\tau$ is the temperature parameter, and $\sigma$ is the sigmoid function.
During training, we use hard thresholding in the forward pass while maintaining differentiability through the straight-through estimator:
\begin{equation}
    z_{o,i} = \begin{cases}
        \mathbb{I}[\tilde{z}_{o,i} > 0.5] & \text{(forward)} \\
        \tilde{z}_{o,i} & \text{(backward)}
    \end{cases}
\end{equation}
At inference, gating becomes deterministic, with $z_{o,i} = \mathbb{I}[\sigma(\alpha_{o,i}) > 0.5]$. MoGA computes the output $\mathbf{h}$ as
% \begin{align}
%     \mathbf h & = \mathbf W_0 \mathbf x + \frac{1}{O}\sum_{o=1}^O\Delta \mathbf W_o \mathbf x \\
%      & = \mathbf W_0 \mathbf x + \frac{1}{O}\sum_{o=1}^O \left( \sum_{i=1}^R z_{o,i} \cdot \mathbf b_i \mathbf a_i^\top \right ) \mathbf x ,
% \end{align}
\begin{align}
    \mathbf h_o & = \mathbf W_0 \mathbf x + \frac{1}{T}\sum_{t=1}^{T}\Delta \mathbf W_{o,t} \mathbf x \\
     & = \mathbf W_0 \mathbf x + \frac{1}{T}\sum_{t=1}^{T} \left( \sum_{i=1}^R z_{o,t,i} \cdot \mathbf b_i \mathbf a_i^\top \right ) \mathbf x ,
\end{align}
where $\Delta \mathbf{W}_{o,t} = \sum_{i=1}^R z_{o,t,i} \cdot \mathbf{b}_i \mathbf{a}_i^\top$ is the object-specific adapter for object $o$ computed from its $t$-th object pointer stored in the memory bank, $z_{o,t,i}$ is the $i$-th component of $\mathbf z_{o,t}$, $T$ is the number of stored pointers, $\mathbf x$ is the input feature, and $\mathbf W_0$ is the frozen base weight.
MoGA shares the same $\{\mathbf a_i, \mathbf b_i\}$ across all objects.
This Siamese structure enables efficient object-specific adaptation without parameter duplication.

% where $\Delta \mathbf{W}_o = \sum_{i=1}^R z_{o,i} \cdot \mathbf{b}_i \mathbf{a}_i^\top$ is the object-specific adapter for object $o$, $z_{o,i}$ is the $i$-th component of $\mathbf z_o$, $\mathbf x$ is the input feature, and $\mathbf W_0$ is the frozen base weight.
% MoGA shares the same $\{\mathbf a_i, \mathbf b_i\}$ across all objects.
% This Siamese structure enables efficient object-specific adaptation without parameter duplication.

\paragraph{Integration and Training.}
We integrate MoGA into SAM2's memory attention module, specifically into the linear projections for self-attention (Q, K) and cross-attention (Q).
Each projection has its own selection of (rank-1) components, though the low-rank adapter and the gating module are shared across objects.
Training uses a standard segmentation loss
\begin{equation}
    \mathcal{L}_{\text{total}} = \frac 1 {T \cdot O} \sum_{t=1}^{T} \sum_{o=1}^{O} \mathcal{L}_{\text{seg}}(y_{o,t}, \hat{y}_{o,t}),
\end{equation}
where $\mathcal{L}_{\text{seg}}$ combines focal and dice losses, $\hat{y}_{o,t}$ is the predicted mask for object $o$ at frame $t$ produced with its memory-conditioned adapter $\Delta \mathbf W_o$, and $y_{o,t}$ is the respective ground-truth mask.
Notably, the gating modules that determine which (rank-1) components to activate are not directly supervised.
They learn to select appropriate adaptation paths solely through the segmentation loss, following established practices in mixture-of-experts literature~\cite{fedus2022switch}.

\section{Experiments}

\subsection{Implementation Details}
We adopt the pre-trained SAM2~\cite{ravi2025sam} and freeze all parameters except for the newly introduced MoGA modules and the LayerNorm layers of the image encoder.
MoGA modules are attached to the projection layers of the memory attention (query and key of self-attention, which share one adapter and gating module, and query of cross-attention) and are trained jointly with LayerNorm, following~\cite{qi2022parameter,valizadehaslani2024layernorm,de2023effectiveness,zhao2023tuning}.
The parameters are optimized with AdamW~\cite{loshchilov2018decoupled} with a learning rate of $5{\times}10^{-6}$, weight decay of $0.1$, batch size of $4$, and a maximum of $3$ objects per clip. The adapter rank $R$ is set to 128. We use a temperature $\tau$ of 0.3 for the Gumbel–sigmoid.

\begin{figure*}[t!]
    \centering
    \includegraphics[width=1.0\linewidth]{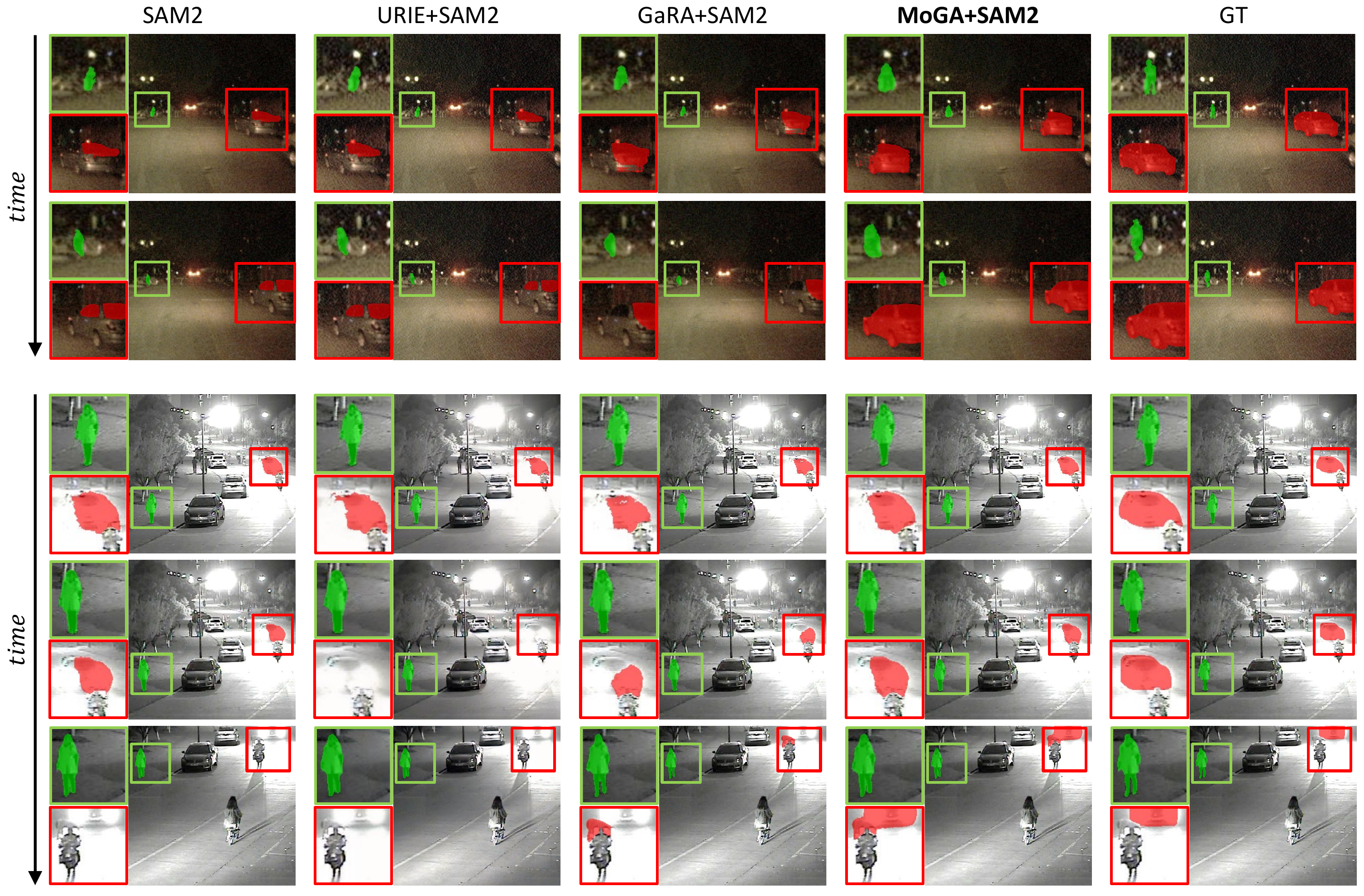}
    \caption{Qualitative results on the real-world corrupted sequences of our benchmark. Each color indicates tracked objects: red for vehicles and green for pedestrians.}
    \label{fig:qual-main}
\end{figure*}
\subsection{Results on Our Real-World Benchmark}

We evaluate each model on our real-world benchmark,  MVSeg-adv~\cite{ji2023mvss} and ACDC-Video~\cite{acdc:v2}, captured under real-world corruptions.
Table~\ref{tab:main_results} reports zero-shot segmentation results.
The original SAM2, despite its strong performance on clear videos, shows substantial degradation under real corruptions, achieving only 69.6\% and 63.5\% $\mathcal{J}\&\mathcal{F}$ on MVSeg-adv and ACDC-Video, respectively.
Restoration methods (URIE+SAM2, AirNet+SAM2) offer marginal improvements and even degrade performance.
Frame-wise application of GaRA similarly achieves small improvements or degrades performance, reaching 69.7\% and 61.3\% $\mathcal{J}\&\mathcal{F}$ on MVSeg-adv and ACDC-Video.
By contrast, MoGA combined with SAM2 (MoGA+SAM2)
achieves significant gains across all metrics, reaching 71.8\% $\mathcal{J}\&\mathcal{F}$ on MVSeg-adv and 64.5\% $\mathcal{J}\&\mathcal{F}$ on ACDC-Video.
This demonstrates the effectiveness of memory-object-conditioned robustification over frame-level approaches.
The improvements are consistent across both datasets, suggesting that MoGA generalizes well to diverse real-world corruption types without dataset-specific tuning.

\subsection{Results on Synthetic Datasets}

We additionally evaluate each model on synthetic corruption datasets, especially on the validation split of YouTube-VOS-C.
Table~\ref{tab:main_results} shows that SAM2 achieves 78.7\% $\mathcal{J}\&\mathcal{F}$ on YouTube-VOS-C, exhibiting a performance degradation from its clear performance, 82.2\%.
Frame-wise robustification methods, including restoration approaches and GaRA, show performance deterioration.
MoGA+SAM2 outperforms all baselines, achieving 79.9\% $\mathcal{J}\&\mathcal{F}$ on corrupted videos.
On clean videos, MoGA+SAM2 reaches 82.6\%, on par with the original SAM2 (82.2\%).
These results suggest that memory-object-based conditioning is more robust compared to frame-wise methods while maintaining reasonable performance on clear videos.

\begin{table}[t]
    \centering
    \caption{Comparison with fully fine-tuned SAM2 on MVSeg-adv.}
\vspace{-2mm}
\begin{tabular}{lccccc}
        \toprule
        Method
        & \makecell{Learnable \\ Params}
        & \makecell{GPU \\ Memory}
        & $\mathcal J \& \mathcal F$\\
        \midrule
        Fine-tuned SAM2      &  80.9M & 25GB  &  71.5   \\
        MoGA+SAM2   & 1.1M & 22GB    & \textbf{71.8} &  \\
        \bottomrule
    \end{tabular}\label{tab:fullft}
\end{table}

\subsection{Qualitative Results}

Figure~\ref{fig:qual-main} presents qualitative results for SAM2, URIE+SAM2, GaRA+SAM2, and MoGA+SAM2 on corrupted real-world videos from our benchmark.
SAM2 predicts noisy masks or loses tracking of objects entirely.
The image restoration method, URIE+SAM2, shows limited improvements and produces partial masks, capturing only fragments of objects under severe corruptions.
Frame-wise application of GaRA, GaRA+SAM2, exhibits temporally inconsistent masks across frames due to its lack of temporal consistency.
In contrast, MoGA+SAM2 shows robust predictions across corrupted frames.
MoGA maintains temporal consistency by leveraging stored object memory, preventing inconsistent results.

\begin{table}[t]
    \centering
    \begin{minipage}[t]{0.51\linewidth}
        \centering
        \caption{ Comparison with LoRA.
    }
\vspace{-2mm}
        \renewcommand{\arraystretch}{1.03}
    \scalebox{0.851}{
        \begin{tabular}{lc}
            \toprule
            Method & $\mathcal{J}\&\mathcal{F}$ \\
            \midrule
            SAM2~\cite{ravi2025sam}        & 69.6 \\
            LoRA~\cite{hu2022lora}+SAM2    & 70.9 \\
            MoGA+SAM2 (Ours)               & \textbf{71.8} \\
            \bottomrule
        \end{tabular}}\label{tab:lora}
    \end{minipage}%
    \hfill
    \begin{minipage}[t]{0.451\linewidth}
        \centering
        \caption{Results on short vs.\ long videos.}
\vspace{-2mm}
        \renewcommand{\arraystretch}{1.1}
    \scalebox{0.8}{
        \begin{tabular}{lcc}
            \toprule
             & SAM2 & Ours \\
            \midrule
            Short (${\sim}$6s)  & 69.6 & \textbf{71.8} \\
            Long (${\sim}$42s)  & 52.3 & \textbf{56.2} \\
            \bottomrule
        \end{tabular}}\label{tab:long}
    \end{minipage}
\end{table}

\begin{figure}[t!]
    \centering
    \includegraphics[width=\linewidth]{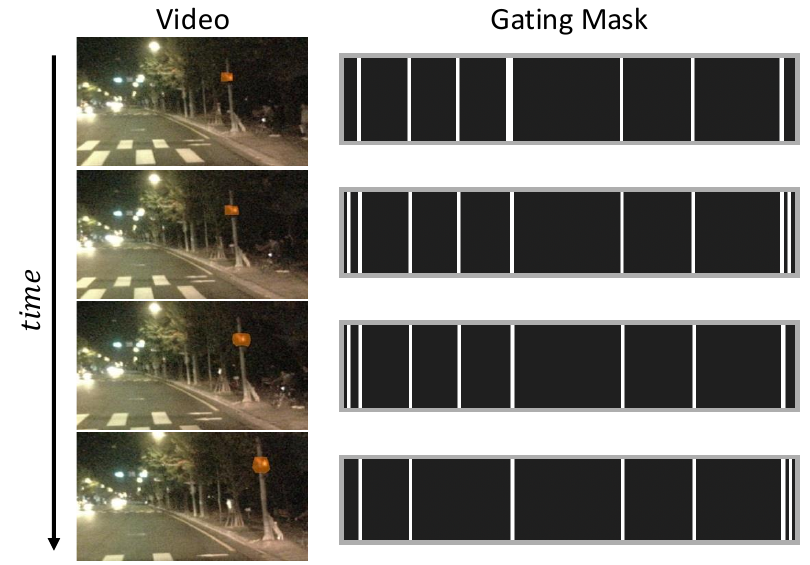}
    \vspace{-2mm}
    \caption{Visualization of gating masks over time. \emph{Left}: input video frames from a nighttime driving sequence. \emph{Right}: corresponding binary gating masks showing activated components (black) for tracked objects.}
    \label{fig:gating}
\end{figure}
\begin{figure}[t!]
    \centering
    \includegraphics[width=0.93\linewidth]{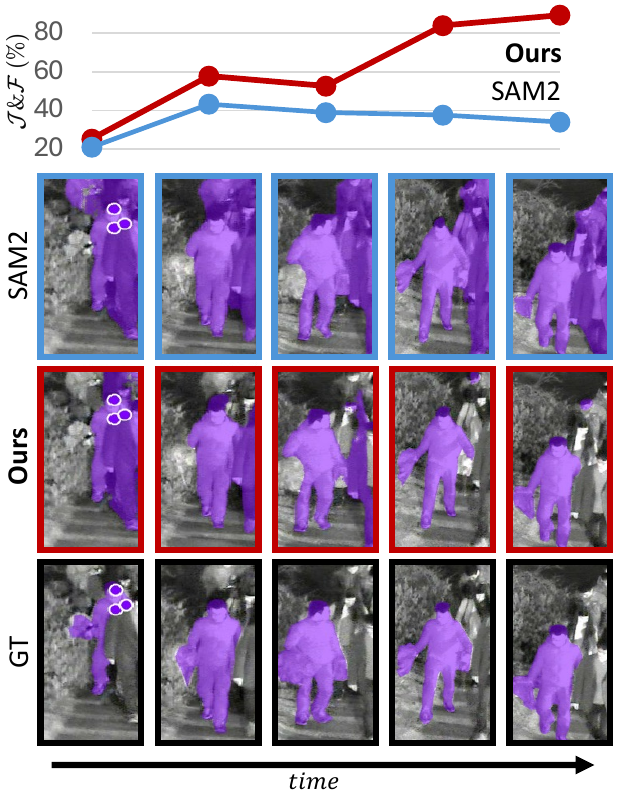}
\vspace{-2mm}
    \caption{Quantitative and qualitative results during inference. \emph{Top}: $\mathcal{J}\&\mathcal{F}$ scores across real-world nighttime frames. \emph{Bottom}: qualitative comparison of segmentation results in the sequence.}
    \label{fig:temporal_adaptation}
\end{figure}
\subsection{Comparison with Full Fine-tuning}
To evaluate parameter efficiency, we compare MoGA combined with SAM2 to a fully fine-tuned SAM2 in Table~\ref{tab:fullft}.
The full fine-tuning baseline optimizes all 80.9M parameters of SAM2, achieving 71.5\% $\mathcal{J}\&\mathcal{F}$ on MVSeg-adv while requiring 25GB of GPU memory for training.
In contrast, MoGA+SAM2 trains only 1.1M parameters yet achieves 71.8\% $\mathcal{J}\&\mathcal{F}$ using only 22GB of memory.
This demonstrates that our memory-conditioned robustification is not only more parameter-efficient but also more effective than full fine-tuning of SAM2.

\subsection{Comparison with LoRA.}
We compare MoGA against LoRA~\cite{hu2022lora} applied to SAM2 under the same training setup.
As shown in Table~\ref{tab:lora}, SAM2+LoRA achieves 70.9\% $\mathcal{J}\&\mathcal{F}$ on MVSeg-adv, improving over the SAM2 baseline (69.6\%) but underperforming MoGA+SAM2 (71.8\%).
LoRA updates weights uniformly, without any awareness of which object is being tracked or what the memory bank contains, applying identical adaptation.
MoGA surpasses LoRA through its memory-object-conditioned gating, which enables temporally consistent and object-specific adaptation essential for RobustPVOS but absent in LoRA.

\begin{table*}[t]
\centering
\setlength{\tabcolsep}{4pt}
\begin{minipage}[t]{0.32\linewidth}
\centering
\caption{Effect of individual components of MoGA on performance on MVSeg-adv.}
    \label{tab:ablate_temporal_object}
    \renewcommand{\arraystretch}{0.9}
    \begin{tabular}{ccc}
        \toprule
        Memory-cond & Object-cond & $\mathcal J \& \mathcal F$ \\
        \midrule
         &  & 69.6 \\
        \cmark &  & 70.9 \\
        \cmark & \cmark & \textbf{71.8} \\
        \bottomrule
    \end{tabular}
\end{minipage}
\hspace{0.01\linewidth}
\begin{minipage}[t]{0.34\linewidth}
\centering
\caption{Effect of adapter rank $R$ of MoGA on performance on YouTube-VOS-C.}
    \label{tab:ablate_rank}
    \renewcommand{\arraystretch}{1.07}
    \setlength{\tabcolsep}{4pt}
    \begin{tabular}{lccccc}
        \toprule
        & \multicolumn{5}{c}{Rank $R$} \\
        \cmidrule(lr){2-6}
        Metric & 32 & 64 & \textbf{128} & 256 & 512\\
        \midrule
        $\mathcal J \& \mathcal F$ & 79.3  & 79.4 & \textbf{79.9} & 79.8 & 79.7  \\
        \bottomrule
    \end{tabular}
\end{minipage}
\hspace{0.01\linewidth}
\begin{minipage}[t]{0.27\linewidth}
\centering
\caption{Effect of temperature $\tau$ of MoGA on YouTube-VOS-C.}
\label{tab:ablate_temp}
\setlength{\tabcolsep}{4pt}
\renewcommand{\arraystretch}{1.07}
\begin{tabular}{lcccc}
    \toprule
    & \multicolumn{4}{c}{Temperature $\tau$} \\
    \cmidrule(lr){2-5}
    Metric &  0.1 & 0.3 & 0.5 & 0.7 \\
    \midrule
    $\mathcal J \& \mathcal F$ &  \textbf{79.9} & \textbf{79.9} & 79.7 & 79.8 \\
    \bottomrule
\end{tabular}\label{tab:temp}
\end{minipage}
\end{table*}

\subsection{In-Depth Analysis}\label{sec:analysis}

\paragraph{Results on Long Videos.}
To assess scalability to longer sequences, we construct extended videos by merging clips (${\sim}6$s each) from the same video into longer sequences (${\sim}42$s, 1K frames) on MVSeg-adv.
As shown in the Table~\ref{tab:long}, SAM2 suffers a notable drop on longer videos.
MoGA+SAM2, however, maintains improvement over SAM2 on both short and long video settings, suggesting that memory-conditioned gating remains stable as the memory bank grows.

\paragraph{Impact of Gating Mask.}
We investigate the effect of memory-conditioned gating by visualizing the learned gating masks across frames in Figure~\ref{fig:gating}.
The visualization reveals that gating masks maintain temporal consistency across the video sequence.
Despite this overall consistency, we observe small adaptive changes in the activated (rank-1) components over time.
These smooth transitions in gating patterns demonstrate that our approach successfully leverages temporal information for stable robustification.

\paragraph{Progressive Improvement of MoGA.}
\Fig{temporal_adaptation} shows quantitative and qualitative comparisons between SAM2 and MoGA+SAM2 model on a sample sequence.
While SAM2 performs around 40\% $\mathcal{J}\&\mathcal{F}$ across the sequence, MoGA progressively improves performance to over 80\% $\mathcal{J}\&\mathcal{F}$ by the end of the sequence.
This behavior arises from the accumulation of object pointers in the memory bank, enabling the gating module to select progressively more suitable components of the adapter.
The qualitative results also demonstrate that MoGA evolves from fragmented masks to complete segmentation, demonstrating the effectiveness of memory-conditioned progress during inference.

\paragraph{Impact of Memory-object Conditioning.}
Table~\ref{tab:ablate_temporal_object} investigates the contribution of each component of MoGA on MVSeg-adv.
The three variants differ as follows: the first uses no conditioning at all; the second applies memory-conditioning only, which aggregates all object pointers into a single shared representation before gating; the third is the full MoGA, which combines memory- and object-conditioning so that gating is performed independently for each object pointer.
Without either form of conditioning, the model achieves only 69.6\% $\mathcal{J}\&\mathcal{F}$.
Incorporating temporal information via memory-conditioning improves performance to 70.9\% $\mathcal{J}\&\mathcal{F}$, demonstrating the importance of object memory for maintaining cross-frame consistency.
The full MoGA, which additionally conditions gating on each individual object pointer, achieves 71.8\% $\mathcal{J}\&\mathcal{F}$, validating the effect of object-specific adaptation on top of temporal conditioning.

\paragraph{Analysis on Rank of Adapter.}
\Tbl{ablate_rank} shows the effect of adapter rank $R$ on YouTube-VOS-C.
The performance remains stable across different rank values, with our default setting of $R=128$ achieving optimal performance at 79.9\% $\mathcal{J}\&\mathcal{F}$.
Lower ranks ($R=32, 64$) yield reasonable performance, showing that compact representations can capture object-specific patterns.
Increasing the rank to 256 or 512 yields a degradation, indicating saturation beyond $R=128$.
The optimal rank of 128 strikes a balance between adaptation flexibility and parameter efficiency, introducing only 1.1M trainable parameters while achieving substantial improvements in robustness.

\paragraph{Analysis on Temperature of Gumbel-Sigmoid.}
\Tbl{ablate_temp} shows the impact of temperature $\tau$ of Gumbel-Sigmoid in \Eq{logit}.
Lower values of $\tau$ produce harder component selection, while higher values yield softer activations.
We find that MoGA is insensitive to $\tau$, with $0.1$ and $0.3$ yielding the best performance on YouTube-VOS-C. Thus, our memory-based conditioning provides strong guidance for gating decisions, enabling hard selection without training instability.

\section{Conclusion}
We have presented the first comprehensive study on robust promptable video object segmentation, establishing it as a central research direction for real-world video object segmentation.
Our first contribution is a carefully designed benchmark comprising real-world evaluation datasets with dense object-level annotations under real adverse conditions and synthetic datasets with controlled temporal degradations.
We have further proposed Memory-object-conditioned Gated-rank Adaptation (MoGA) as an effective baseline that leverages object representations from the memory bank to achieve object-specific robustification while maintaining temporal consistency across frames.
Extensive experiments demonstrate that memory-based conditioning substantially outperforms per-frame approaches, validating its effectiveness for handling spatially and temporally varying degradations in videos.
Our comprehensive benchmark and baseline provide foundations for future research in RobustPVOS.

\vspace{3mm}
{\small
\noindent \textbf{Acknowledgement.}
This work was supported by AI Graduate School
Program at POSTECH (RS-2019-II191906), and the IITP grants (RS-2024-00457882, RS-2022-II220926, RS-2024-00509258, RS-2024-00469482) funded by the Ministry of Science and ICT, Korea.
}

\clearpage
{
    \small
    \bibliographystyle{ieeenat_fullname}
    \bibliography{main}
}

\end{document}